\documentclass[sigconf]{acmart}
\AtBeginDocument{%
  }

\theoremstyle{plain}

\theoremstyle{definition}
\newtheorem{definition}{Definition}

\theoremstyle{remark}

\usepackage{multirow}
\usepackage{subfigure}
\usepackage{xcolor}

\acmYear{2025}

\usepackage{todonotes}

\setcopyright{none}
\begin{document}
\renewcommand\footnotetextcopyrightpermission[1]{}
\title{UM\textsuperscript{3}: Unsupervised Map to Map Matching}

\author{Chaolong Ying}
\email{chaolongying@link.cuhk.edu.cn}
\affiliation{%
  \institution{The Chinese University of Hong Kong, Shenzhen}
  \city{Shenzhen}
  \state{Guangdong}
  \country{China}
}

\author{Yinan Zhang}
\email{yinanzhang1@link.cuhk.edu.cn}
\affiliation{%
  \institution{The Chinese University of Hong Kong, Shenzhen}
  \city{Shenzhen}
  \state{Guangdong}
  \country{China}
}

\author{Lei Zhang}
\email{leizhang1@link.cuhk.edu.cn}
\affiliation{%
  \institution{The Chinese University of Hong Kong, Shenzhen}
  \city{Shenzhen}
  \state{Guangdong}
  \country{China}
}

\author{Jiazhuang Wang}
\email{wangjiazhuang@mxnavi.com}
\affiliation{%
  \institution{MXNavi Co.,Ltd.}
  \city{Shenyang}
  \state{Liaoning}
  \country{China}
}

\author{Shujun Jia}
\authornotemark[1]
\email{jiashujun@mxnavi.com}
\affiliation{%
  \institution{MXNavi Co.,Ltd.}
  \city{Shenyang}
  \state{Liaoning}
  \country{China}
}

\author{Tianshu Yu}
\authornote{Corresponding authors.}
\email{yutianshu@cuhk.edu.cn}
\affiliation{%
  \institution{The Chinese University of Hong Kong, Shenzhen}
  \city{Shenzhen}
  \state{Guangdong}
  \country{China}
}








\renewcommand{\shortauthors}{Ying et al.}

\begin{abstract}
Map-to-map matching is a critical task for aligning spatial data across heterogeneous sources, yet it remains challenging due to the lack of ground truth correspondences, sparse node features, and scalability demands. In this paper, we propose an unsupervised graph-based framework that addresses these challenges through three key innovations. First, our method is an unsupervised learning approach that requires no training data, which is crucial for large-scale map data where obtaining labeled training samples is challenging. Second, we introduce pseudo coordinates that capture the relative spatial layout of nodes within each map, which enhances feature discriminability and enables scale-invariant learning. Third, we design an mechanism to adaptively balance feature and geometric similarity, as well as a geometric-consistent loss function, ensuring robustness to noisy or incomplete coordinate data. At the implementation level, to handle large-scale maps, we develop a tile-based post-processing pipeline with overlapping regions and majority voting, which enables parallel processing while preserving boundary coherence. Experiments on real-world datasets demonstrate that our method achieves state-of-the-art accuracy in matching tasks, surpassing existing methods by a large margin, particularly in high-noise and large-scale scenarios. Our framework provides a scalable and practical solution for map alignment, offering a robust and efficient alternative to traditional approaches.

\end{abstract}

\maketitle

\section{Introduction}
Maps are fundamental tools for understanding and navigating the spatial complexity of the real world. Over the last decade, advancements in geospatial data acquisition and computational power have led to the proliferation of map-related datasets, ranging from street maps and transportation networks to topographic and environmental maps~\cite{ruiz2011digital}. In this context, the problem of map-to-map matching, which refers to the alignment of maps from different sources or formats to identify similarities and differences between them, has become a critical challenge with far-reaching applications~\cite{aguilar2024graph, faerman2019graph}. For instance, map companies often need to align maps from different time periods to track changes in road networks, ensuring that their datasets reflect the most up-to-date geographical information. This process typically involves matching geographic features on one map with corresponding features on another map, enabling the integration and analysis of spatial information.

Accurate map-to-map matching is fundamental for several  reasons. One key motivation is the integration of diverse geospatial data sources for monitoring changes in geographic features, such as urban expansion or environmental shifts. For instance, governmental cadastral maps provide highly accurate and authoritative depictions of roads and land boundaries but may lack the frequent updates characteristic of crowd-sourced maps like OpenStreetMap (OSM)~\cite{haklay2008openstreetmap}. Aligning these complementary datasets allows planners and researchers to combine timely updates with high-precision ground-truth information, benefiting applications such as infrastructure development and urban analytics~\cite{goodchild2012assuring}.
Another significant requirement arises in autonomous vehicle navigation. Self-driving systems often depend on high-definition maps for lane-level navigation, which must be reconciled with baseline road maps to ensure accuracy at scale. Conflicting geometries or naming conventions between maps can result in inconsistent path planning or create safety risks~\cite{wong2020mapping}.

Map-to-map matching, a crucial task in geospatial data integration, faces several key challenges that complicate its implementation. These include data heterogeneity, feature inconsistency, noise-induced errors, computational complexity, and the reliance on manual annotation for labeled data. Maps from different sources often vary in coordinate systems, scales, and feature representations, making direct alignment difficult. Feature inconsistency arises when the same geographic entity is represented differently across maps, such as variations in naming, geometry, or status. Noise from data collection processes, like GPS inaccuracies or digitization errors, further degrades matching accuracy, especially in dense urban environments~\cite{soni2022finding}. The computational complexity of processing large, dynamic datasets with intricate topological relationships demands efficient yet precise algorithms~\cite{zeidan2020geomatch,ying2024boosting}. Additionally, the lack of automated methods for generating high-quality labeled data means that annotations often rely on manual effort, which is time-consuming, costly, and prone to human error. Addressing these challenges necessitates advanced data processing, innovative algorithms, and domain expertise to achieve reliable and scalable map-to-map matching solutions.

Traditional approaches to map-to-map matching have primarily relied on geometric, topological, and rule-based methods to align and correlate features across different maps. Geometric methods focus on spatial similarity, using metrics such as Euclidean distance or shape similarity to match corresponding features based on their spatial proximity and geometric properties~\cite{chehreghan2018geometric,wang2021solving}. While effective in simple scenarios, these methods often struggle with data heterogeneity and noise, as they do not account for contextual or semantic relationships. Topological methods, on the other hand, emphasize the relationships between features, such as connectivity and adjacency, to improve matching accuracy by analyzing the structure of road networks or the arrangement of geographic entities~\cite{mustiere2008matching}. However, they require robust preprocessing to ensure consistent topological representations across maps, which can be challenging with incomplete or inconsistent data. Rule-based methods incorporate domain knowledge and heuristic rules, such as matching features with similar names or classifications, to guide the matching process~\cite{walter1999matching}. While these methods can improve accuracy in specific contexts, they are often limited by the quality and generality of the rules, making them less adaptable to diverse datasets or dynamic environments. Despite their strengths, traditional approaches face limitations in scalability, robustness, and automation, often requiring significant manual intervention for parameter tuning, rule definition, or error correction~\cite{li2011optimisation}. As a result, recent research has increasingly turned to data-driven and machine learning-based methods to address these shortcomings and enhance the performance of map-to-map matching systems.

To the best of our knowledge, there is currently no machine learning-based method specifically designed for map-to-map matching. This gap can be attributed to several significant challenges. First, obtaining large-scale, high-quality map datasets for training is inherently difficult due to the scarcity of publicly available map data and the complexity of integrating heterogeneous map sources. Second, even if such datasets were available, annotating matching information between maps requires extensive domain expertise from professional map engineers, making the labeling process both time-consuming and costly. These challenges collectively render traditional supervised learning approaches impractical for large-scale map-to-map matching applications. To address these limitations, we propose an innovative approach named Unsupervised Map-to-Map Matching ($\mathrm{UM}\textsuperscript{3}$) that leverages the inherent structural and topological characteristics of maps to achieve accurate matching without the need for manually labeled data. By formulating the problem as an optimization task, our method enables the system to iteratively learn and refine matching patterns, delivering precise alignment results for previously unseen maps in a computationally efficient manner. This approach not only eliminates the dependency on labeled data but also ensures high accuracy and scalability, making it well-suited for real-world geospatial applications. We summarize our contributions as follows:
\begin{itemize}
    \item \textbf{Unsupervised Approach.} We propose a novel unsupervised map-to-map matching method that eliminates the need for manually labeled data, addressing the resource-intensive and time-consuming limitations of traditional supervised learning approaches.
    \item \textbf{Optimization-Based Framework.} By modeling map-to-map matching as an optimization task, our method iteratively learns and refines matching patterns, enabling accurate alignment of large-scale maps without prior knowledge or human intervention.
    \item \textbf{Scalability and Efficiency.} Our approach is designed to handle large-scale geospatial datasets efficiently, providing a scalable solution that achieves high accuracy while significantly reducing computational and manual resource requirements.
\end{itemize}

\section{Related Work}
\paragraph{Point Cloud Matching}
Point cloud matching focuses on aligning 3D point sets, often captured by LiDAR or other sensors, and has been widely studied in robotics, autonomous driving, and 3D reconstruction. Traditional methods, such as the Iterative Closest Point (ICP) algorithm~\cite{besl1992method}, iteratively minimize the distance between corresponding points but struggle with noise, outliers, and partial overlaps, making it prone to falling into incorrect local optima. 
An alternative approach to point cloud matching is based on optimal transport theory, such as Sinkhorn Distance (SD)~\cite{cuturi2013sinkhorn}. It treats point clouds as probability distributions, computing an optimal transport plan between these distributions and aiming to minimize the transportation cost. Moreover, the method introduces entropy regularization, which smooths the classical optimal transportation problem. The Sinkhorn-Knopp algorithm ~\cite{knight2008sinkhorn} enables fast convergence, solving the regularized problem more efficiently.
Recent work has explored the integration of point cloud matching into the deep learning framework.  One prominent example is PointNetLK~\cite{aoki2019pointnetlk}, which combines the feature extraction capabilities of PointNet~\cite{qi2017pointnet} with the optimization framework of the Lucas-Kanade ~\cite{baker2004lucas} optical flow method. PointNetLK directly learns the rigid transformation between point clouds by iteratively refining the alignment, using a learned feature representation to guide the transformation process. Another notable approach is Deep Closest Point (DCP)~\cite{wang2019deep}
. It employs PointNet ~\cite{qi2017pointnet} and DGCNN ~\cite{phan2018dgcnn} for feature embedding to capture point cloud representations, Transformer for the attention mechanism to refine correspondences, and SVD for computing the rigid transformation, which enables precise alignment of the point clouds.

\paragraph{Graph Matching.} Graph matching aims to find correspondences between nodes and edges in two graphs, with applications in computer vision, bioinformatics, and network analysis. Early approaches, such as spectral matching~\cite{leordeanu2005spectral}, use graph Laplacians to identify correspondences but are sensitive to noise and structural differences. More recent methods incorporate deep learning to extract more dedicated features and optimize matching objectives~\cite{zanfir2018deep,yu2019learning,sarlin2020superglue}. Other efforts have been devoted to designing advanced pipelines, such as combining with traditional non-differentiable combinatorial solvers~\cite{rolinek2020deep}, converting the graph matching problem into node classification problem on the association graphs~\cite{wang2021neural}, and learning latent topology to enhance matching~\cite{yu2021deep}. While these techniques have shown promise, they often rely on supervised learning and struggle with scalability when applied to large graphs or heterogeneous data.

\paragraph{Trajectory-Map Matching.} Trajectory-map matching involves aligning GPS trajectories to a digital road network, a critical task for navigation, traffic analysis, and location-based services. Early methods, such as geometric approaches~\cite{white2000some}, match trajectories to the nearest road segments based on spatial proximity but often fail to account for topological constraints. Advanced techniques, like Hidden Markov Models (HMM)~\cite{newson2009hidden}, incorporate probabilistic frameworks to improve accuracy by considering both spatial and temporal information. More recent work leverages machine learning to enhance matching performance, such as using deep neural networks to learn trajectory-road relationships~\cite{liu2023graphmm}. They employ spatial-temporal encoders to extract the semantics of trajectories and predict their correspondences with maps during the inference stage~\cite{ren2021mtrajrec,wei2024micro}. These methods typically require extensive labeled trajectory data and are computationally demanding for large-scale or real-time applications. While trajectory-map matching shares some common ground with map-to-map matching, the two tasks are fundamentally different. In map-to-map matching, the goal is to find correspondences between two maps by aligning nodes and edges based on structural similarity. In contrast, trajectory-map matching involves a dynamic, time-dependent alignment, where the challenge lies in associating spatiotemporal data.

\section{Preliminaries}
We briefly review the background of this topic in this section, as well as elaborate on the notations.
\subsection{Problem Definition}
\begin{definition}
\textbf{Map.} A map $M$ is defined as an undirected graph $G = (\mathcal{V}, \mathcal{E})$ with $n$ nodes, where $\mathcal{V}$ is the set of nodes and $\mathcal{E}$ is the set of edges. Each node $v \in \mathcal{V}$ corresponds to a geographical location defined by its latitude and longitude coordinates $\mathbf{x}$, while each edge $e \in \mathcal{E}$ represents a connection between two nodes, typically corresponding to a segment of a road. It is also practical to represent the graph with an adjacency matrix $\mathbf{A} \in \{0, 1\}^{n \times n}$. 


A map dataset typically consists of a collection of roads, where each road is a sequence of nodes that form a continuous path. Formally, a road $R$ can be defined as an ordered sequence of nodes $R = (v_1, v_2, \dots, v_k)$, where $v_i \in \mathcal{V} $ for $ i = 1, 2, \dots, k$, and each pair of consecutive nodes $ (v_i, v_{i+1}) $ is connected by an edge $ e_i \in \mathcal{E} $. Each road is associated with a unique road ID, which serves as an identifier for the road within the map.
\end{definition}

\begin{definition}
\textbf{Map-to-Map Matching.} Given source map $M_s$ and target map $M_t$, the map-to-map matching problem aims to establish a correspondence between the roads in $M_s$ and $M_t$ such that the alignment maximizes spatial consistency, which refers to ensuring that corresponding roads in the two maps are matched based on their geometric properties and relative positions.
\end{definition}

\subsection{Graph Neural Networks}
Graph Neural Networks (GNNs) are a class of deep learning models specifically designed to process data represented as graphs. By leveraging the graph structure, GNNs can capture complex dependencies and interactions, enabling them to learn rich representations of nodes and entire graphs. These models utilize message passing mechanisms to iteratively aggregate information from neighboring nodes, ultimately allowing them to excel in various applications. As GNNs continue to evolve, they offer powerful tools for tackling challenges in machine learning and artificial intelligence across diverse domains. Given an input graph, typical GNNs compute node embeddings $\boldsymbol{h}_u^{(t)} , \forall u \in \mathcal{V}$ with $T$ layers of iterative message passing~\cite{xu2018powerful}:
\begin{equation}
    \boldsymbol{h}_u^{(t+1)} = \psi \left(\boldsymbol{h}_u^{(t)}, \sum_{v\in \mathcal{N}_u }  \boldsymbol{h}_v^{(t)}\cdot \phi (\boldsymbol{e}_{uv}) \right)
\end{equation}
for each $t \in [0, T-1]$, where $\mathcal{N}_u = \{v\in {\mathcal{V}} | (u, v) \in {\mathcal{E}}\}$, while $\psi$ and $\phi$ are neural networks, e.g. implemented using multilayer perceptrons (MLPs).

\subsection{Graph Matching}
Graph matching is a fundamental problem in computer science and related fields~\cite{yu2019learning, wang2021neural, ying2025neural}, aiming to establish correspondences between the nodes and edges of two graphs. Given two graphs $G_1$ and $G_2$, the goal is to find a mapping $f: \mathcal{V}_s \rightarrow \mathcal{V}_t$ that maximizes both structural and attribute consistency between the graphs. Recent works often model graph matching as Lawler's Quadratic Assignment Problem (QAP)~\cite{loiola2007qap}, a well-known combinatorial optimization problem. In this formulation, the goal of graph matching problem is to find a hard correspondence matrix $\mathbf{P} \in \{0, 1\}^{n_s \times n_t}$ that maximizes a compatibility function while adhering to row and column constraints. This optimization problem is expressed as:
\begin{equation}
\begin{array}{ll} 
    \max_{\mathbf{P}} &\mathrm{vec}(\mathbf{P})^\top \mathbf{K}\mathrm{vec}(\mathbf{P})\\
    \text { s.t. } & \mathbf{P} \in\{0,1\}^{n_s \times n_t}, \mathbf{P} \mathbf{1}_{n_t}=\mathbf{1}_{n_s}, \mathbf{P}^{\top} \mathbf{1}_{n_s} \leq \mathbf{1}_{n_t},
\end{array}
\label{eq: QAP}
\end{equation}
where $\mathbf{K}$ is the affinity matrix derived from the structural information of $G_s$ and $G_t$ and $\mathbf{1}_{n}$ is a column vector of length $n$ whose elements are all equal to 1. The hard correspondence matrix  can be relaxed into a double-stochastic matrix $\mathbf{S} \in [0,1]^{n_s \times n_t}$. In this formulation, each row of 
$\mathbf{S}$ sums to 1, and each column sums to a value 
$\leq 1$, ensuring that the correspondence probabilities are well-defined. 

\begin{figure*}[tb]
    \centering
    \includegraphics[width=0.7\textwidth]{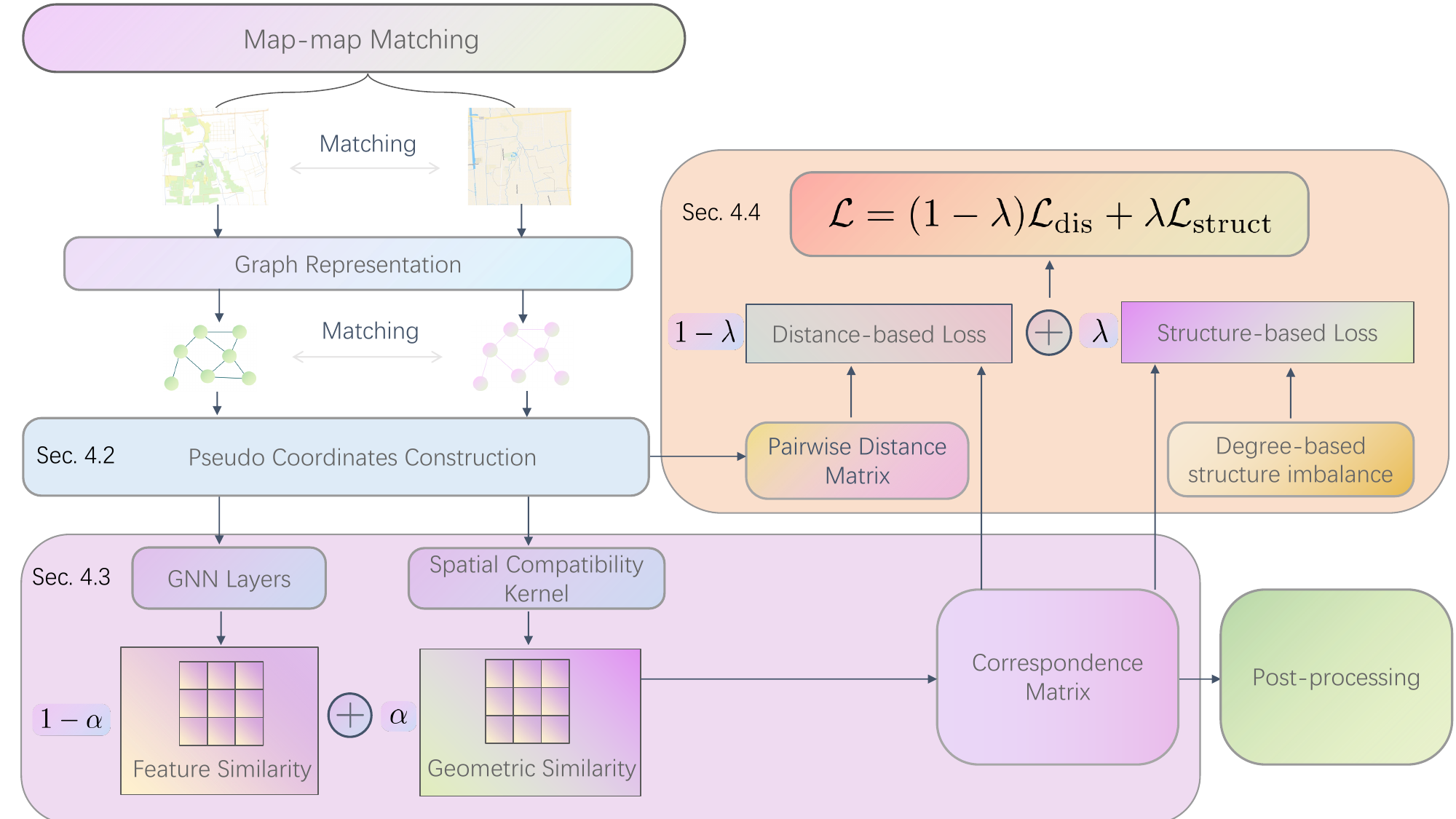}
    \caption{Overview of Unsupervised Map-to-Map Matching. The key steps include: (1) transforming raw GPS coordinates into pseudo coordinates that preserve spatial topology; (2) learning a correspondence matrix through the fusion of feature and geometric similarity; and (3) optimizing a map-to-map matching–specific loss that incorporates spatial and structural constraints. These steps correspond to Sections~\ref{sec: pseudo coordinates construction}, ~\ref{sec: learning node corespondence}, and~\ref{sec: unsupervised loss function}, respectively.}
    \label{fig: overview}
\end{figure*}

\section{Methodology}
\subsection{Overview}
Map-to-map matching is a challenging problem that seeks to find correspondences between two maps represented as graphs. Unlike traditional graph-graph matching, which often relies on supervised learning with known node-node correspondences as ground truth, our problem setting is fundamentally unsupervised: no ground truth correspondences are available in map data. Moreover, while image-based graph matching benefits from rich node features (e.g., pixel intensities or deep features), map data provides limited discriminability of node features—primarily latitude and longitude coordinates. Combined with the significant variability in road structures and coordinate scales across different regions, it is thus difficult to directly apply existing graph matching techniques.

To address these challenges, we propose an unsupervised map-to-map matching framework that formulates the matching problem as an optimization task. Our approach does not rely on labeled correspondences but instead learns to align maps by optimizing a carefully designed objective function. The key steps of our methodology are as follows: (1) transforming raw latitude and longitude coordinates into pseudo coordinates, a normalized and topology-aware representation that captures the relative spatial structure of the map; (2) extracting high-level node embeddings using a GNN $\Psi_\theta(\cdot)$ in Eq.~\ref{eq:feat_gnn} and learning a correspondence matrix that aligns the nodes of the two maps by the fusion of feature similarity and geometric similarity in Eq.~\ref{eq: similarity fusion}; and (3) designing a novel loss function tailored to the characteristics of map-to-map matching, which incorporates both spatial and structural constraints to guide the optimization process. An overview of our proposed framework is illustrated in Figure~\ref{fig: overview}. By combining these steps, our method achieves robust and accurate map-to-map matching without the need for supervised training data.


\subsection{Pseudo Coordinates Construction}
\label{sec: pseudo coordinates construction}
To address the challenges of limited discriminability and the inherent variability in geographical coordinates across different regions, we propose a novel pseudo coordinates construction method that transforms raw latitude and longitude values into the relative spatial layout of nodes within each map. This section details the key steps and motivations behind our approach. 

For each node $v$, we denote its node features as $\mathbf{x} = [x^{(1)}, x^{(2)}] = [\mathrm{lat}(v), \mathrm{lon}(v)]$, where $\mathrm{lat}(\cdot)$ and $\mathrm{lon}(\cdot)$ are the latitude and longitude values of a node. We first compute the minimum and maximum latitude and longitude values across all nodes in the map:
\begin{equation}
\label{eq: min max}
\begin{aligned}
    \mathrm{lat}_{\mathrm{min}} &= \min_{v \in \mathcal{V}} \mathrm{lat}(v), \quad
    \mathrm{lat}_{\mathrm{max}} = \max_{v \in \mathcal{V}} \mathrm{lat}(v),
\\
    \mathrm{lon}_{\mathrm{min}} &= \min_{v \in \mathcal{V}} \mathrm{lon}(v), \quad
    \mathrm{lon}_{\mathrm{max}} = \max_{v \in \mathcal{V}} \mathrm{lon}(v).
\end{aligned}
\end{equation}
These bounds define the spatial envelope of the map and serve as a reference frame for normalization. Then for each node $v \in \mathcal{V}$, we compute its pseudo coordinates $(\bar{x}^{(1)}, \bar{x}^{(2)})$ by normalizing its raw coordinates within the global bounds:
\begin{equation}
\label{eq: normalize}
\begin{aligned}
 \bar{x}^{(1)} &= \frac{\text{lat}(v) - \text{lat}_{\text{min}}}{\text{lat}_{\text{max}} - \text{lat}_{\text{min}}}, \qquad
   \bar{x}^{(2)}  = \frac{\text{lon}(v) - \text{lon}_{\text{min}}}{\text{lon}_{\text{max}} - \text{lon}_{\text{min}}}
\end{aligned}.
\end{equation}

By normalizing coordinates within the global bounds, the representation becomes agnostic to the absolute scale of the map, enabling consistent learning across regions. The relative positioning of nodes within the map is explicitly encoded, allowing the model to reason about spatial proximity and connectivity. The pseudo coordinates serve as initial node features for GNNs, providing a structured input that aligns with the inductive biases of message-passing architectures.

\subsection{Learning Node Correspondence}
\label{sec: learning node corespondence}
Given the pseudo coordinates as initial node features, the next step in our framework is to learn a node correspondence matrix that aligns the nodes of two maps. Our approach formulates the map-to-map matching problem as an unsupervised optimization task, where the goal is to find a correspondence matrix that maximizes the similarity between the two graphs. Different from traditional graph matching frameworks that only focus on feature similarity, we propose a geometric-aware similarity fusion mechanism that unifies feature affinity and spatial proximity into a joint optimization framework. This innovation is motivated by the observation that map data inherently exhibits a duality of topological connectivity and geometric continuity, which must be harmonized for robust correspondence learning.

\paragraph{Feature Similarity Learning} 
As for learning the feature similarity,  we follow related approaches~\cite{fey2020deep} to model the node-to-node correspondence by computing pairwise node similarities between the two maps. Specifically, given the initial node coordinate matrix $\mathbf{X}$, the edge feature matrix $\mathbf{E}$ is constructed by computing the great-circle distance between nodes. Together with the pseudo coordinates as initial node features, the latent node embeddings are computed as
\begin{equation}\label{eq:feat_gnn}
\mathbf{H}_s = \Psi_\theta(\mathbf{A}_s, \mathbf{\bar{X}}_s, \mathbf{E}_s) \qquad
\mathbf{H}_t = \Psi_\theta(\mathbf{A}_t, \mathbf{\bar{X}}_t, \mathbf{E}_t)
\end{equation} 
through a shared neural network $\Psi_\theta$. Then feature similarity is obtained by
\begin{equation}
    \hat{\mathbf{S}} = \mathbf{H_s^{\top}\mathbf{H}_t}.
\label{eq: sinkhorn}
\end{equation}
In our implementation, $\Psi_\theta$  is a GNN designed to obtain permutation-equivariant node representations~\cite{hamilton2017representation}.

\paragraph{Geometric Similarity Learning}

When two maps correspond to the same geographic region, their respective nodes should be spatially close to their true counterparts. In an ideal scenario, a node in one map should not be matched to a node that is far away in the other map. However, due to factors such as structural variations, noise, and incomplete data, erroneous matches may still occur.

To mitigate such errors, we introduce a Geometric Similarity Learning module that incorporates spatial constraints into the matching process. This module ensures that the learned correspondences favor geometrically plausible matches by penalizing associations between nodes that are significantly distant from each other. By integrating this geometric prior into the initial feature based similarity, we improve the robustness of our method against incorrect matches and enhance the overall accuracy of the map alignment process. Specifically, we introduce a spatial compatibility kernel $\mathrm{exp}(-\psi_\theta(\mathbf{D}))$, derived from the pairwise distance matrix 
$\mathbf{D}$, to refine the initial feature-based similarity matrix. Here $\psi_\theta$ is an MLP, and the pairwise distance matrix $\mathbf{D} \in \mathbb{R}^{n_s \times n_t}$ is computed based on the pseudo coordinates for the two maps, where each entry $\mathbf{D}_{uv}$ represents the Euclidean distance between node $u \in \mathcal{V}_s$ and $v \in \mathcal{V}_t$:
\begin{equation}
    \mathbf{D}_{uv} = ||\mathbf{\bar{X}}_1[u] - \mathbf{\bar{X}}_2[v] ||_2.
\end{equation}
Here $\mathbf{\bar{X}}_s[u]$, $\mathbf{\bar{X}}_t[v]$ denote the pseudo coordinates of nodes $u$ and $v$. Finally, the correspondence matrix is learned by the fusion of feature and geometric similarity: 
\begin{equation}
\label{eq: similarity fusion}
    \mathbf{S} = \mathrm{Sinkhorn}(\alpha\hat{\mathbf{S}} + (1-\alpha)\mathrm{exp}(-\psi_\theta(\mathbf{D}))),
\end{equation}
where $\mathrm{Sinkhorn}(\cdot)$ is a normalization operator to obtain double-stochastic correspondence matrices~\cite{sinkhorn1967concerning}, $\alpha$ is a \textbf{learnable} hyperparameter balancing the contribution of feature similarity and geometric similarity, taking into account that in noisy scenarios, the coordinates may be inaccurate, and thus the spatial compatibility kernel may not provide reliable refinements.

The proposed spatial compatibility kernel acts as a spatial attention gate, adaptively amplifying the similarity scores for node pairs that are both feature-similar and geometrically proximate, while suppressing matches that lack spatial coherence. This mimics the human cognitive process of map alignment, where structural consistency and positional continuity are jointly evaluated.

\subsection{Unsupervised Loss Function}
\label{sec: unsupervised loss function}
To guide the learning of node correspondences in an unsupervised manner, we design a novel loss function that leverages the pairwise distances between node in the two maps. The key idea is to encourage the correspondence matrix $\mathbf{S}$ to align nodes that are not only feature-similar but also structurally consistent in terms of their spatial relationships.

 We define $\mathcal{L}_{\mathrm{dis}}$ as the L$_2$ norm of the element-wise product of the pairwise distance matrix $\mathbf{D}$ and the correspondence matrix $\mathbf{S}$:
\begin{equation}
    \mathcal{L}_{\mathrm{dis}} = ||\mathbf{D} \odot \mathbf{S}||_2,
\end{equation}
where $\odot$ denotes the Hadamard product operator.

In real-world scenarios, the accuracy of coordinate information can be compromised due to noise, measurement errors, or incomplete data. Relying solely on coordinate-based features may lead to suboptimal matching results, especially when the noise level is high. To address this issue, we draw inspiration from the irregularity measure in graph theory~\cite{albertson1997irregularity} to minimize the imbalance between graph pairs, which posits that matched nodes should exhibit balanced structural properties in addition to feature and geometric similarity. 

To this end, we propose a structure-based loss term for map-to-map matching task 
$\mathcal{L}_{\mathrm{struct}}$ that penalizes mismatches in the local structural properties of nodes. Specifically, we compute a structural difference matrix $\mathbf{T} \in \mathbb{R}^{n_s \times n_t}$, where each entry $\mathbf{T}_{uv}$ is computed by 
\begin{equation}
    \mathbf{T}_{uv} = \mathrm{Deg}(u) - \mathrm{Deg}(v),
\end{equation}
where $\mathrm{Deg}(u)$ is the normalized degree of node $u$ based on the one-hop neighborhood of a node, defined as:
\begin{equation}
    \mathrm{Deg}(u) = \frac{1}{|\mathcal{N}_u| + 1} \sum_{v \in \mathcal{N}_u \cup \{u\}} \mathrm{degree}(v),
\end{equation}
where $\mathrm{degree}(v)$ is the degree of $v$. Similarly, $\mathcal{L}_{\mathrm{struct}}$ is defined as
\begin{equation}
    \mathcal{L}_{\mathrm{struct}} = ||\mathbf{T} \odot \mathbf{S}||_2.
\end{equation}
To ensure robust and accurate map-to-map matching, we design a composite loss function that combines both distance-based and structure-based constraints. The overall loss function is defined as:
\begin{equation}\label{eq:overall_loss}
    \mathcal{L} = (1-\lambda)\mathcal{L}_{\mathrm{dis}} + \lambda \mathcal{L}_{\mathrm{struct}},
\end{equation}
where $\lambda$ is a weighting hyperparameter that balances the contributions of the two loss terms.

During the inference phase, the hard assignment is obtained via the Hungarian algorithm~\cite{burkard2012assignment} as a post processing step, i.e. $\mathbf{P} = \mathrm{Hungarian}(\mathbf{S})$, where $\mathbf{S}$ is computed via Eq.~\ref{eq: similarity fusion}.

\subsection{Extension to Large-scale Maps}
\label{sec: discussion}


For large-scale maps, processing the entire graph in a single batch is often infeasible due to memory limitations. To address this, we adopt a tile-based processing strategy. The input map is divided into a $k \times k$ grid of overlapping tiles, with adjacent tiles overlapping by a predefined portion of their spatial extent to mitigate information loss at tile boundaries. This overlap ensures that nodes near tile edges are included in multiple tiles, reducing fragmentation of matches across boundaries. Nodes in overlapping regions may receive conflicting correspondence assignments from different tiles. To resolve this, we implement a majority voting scheme, where each node in the overlap region collects all candidate matches from the tiles it belongs to, and the final match is determined by selecting the correspondence with the highest aggregated probability across all overlapping tiles. Ties are resolved by prioritizing matches with smaller spatial distances in the pseudo coordinate space.

Since tiles are processed independently, our framework naturally supports parallel computation. On GPU-enabled systems, multiple tiles can be processed concurrently using batched operations. For distributed systems, tiles can be assigned to different workers, with results aggregated centrally after processing. This design significantly reduces wall-clock time for large-scale maps, ensuring that our method is both practical for real-world applications and scalable to industrial-scale map datasets.

\section{Experiments}
In this section, we evaluate the performance of our method on map-to-map matching problem. We conduct experiments on real-world datasets collected from three distinct geographic regions to validate its effectiveness in practical scenarios. To assess the robustness of our method under noisy conditions, we generate synthetic datasets by introducing controlled levels of noise to the coordinates and evaluate its performance against baseline methods. To comprehensively assess performance, we compare our method against several baseline approaches in terms of both accuracy and computational efficiency, demonstrating its advantages in large-scale map-matching tasks. Finally, we conduct a series of parameter analyses and ablation studies to evaluate the effectiveness of our method.

\subsection{Experimental Setup}

\paragraph{Datasets.} Our experiments are conducted on map data from three different regions: Boston (USA), Ichikawa (Japan), Shanghai (China), and Bremen (Germany), with statistics in Table~\ref{tab:datasets}. The dataset for Boston is obtained from a publicly available source~\cite{BostonOpenData}, the dataset for Ichikawa is acquired from an official government website~\cite{Okada2017UrbanFaultMap}, the dataset for Shanghai is collected from an industrial partner, and the dataset for Bremen is from OSM~\cite{haklay2008openstreetmap}. To evaluate the performance in real-world scenarios, we align these datasets with their corresponding regions in the latest OSM data. Specifically, in the Bremen dataset, we perform map-to-map matching between the 2014 and 2025 maps of a selected region, with both maps collected from OSM. Since the source and target maps originate from the same data source, their identifiers are consistent (e.g., unique road IDs in OSM), which makes it straightforward to obtain ground-truth correspondences. Due to differences in data sources, collection methods, and acquisition times, there exist inherent discrepancies between the collected road network data and the corresponding OSM data. These inherent differences make the dataset particularly challenging and realistic for evaluating map-to-map matching algorithms. The variety in geographic regions and data sources ensures the robustness of our method across different road network structures and data acquisition conditions.

To evaluate the robustness of our method under noisy conditions, we create synthetic datasets by introducing artificial noise into the OSM road networks of the Boston, Ichikawa, and Shanghai regions. The noise model is inspired by prior work~\cite{newson2009hidden}, which suggests that GPS signals can be approximated by Gaussian noise with a mean of 0 and a standard deviation of $\sigma = 4.07$ meters. Based on this model, we define three noise levels to simulate varying degrees of GPS inaccuracies: low noise $\sigma$, medium noise $5\sigma$, and high noise $10\sigma$. For each noise level, we independently perturb the latitude and longitude coordinates of every node in the original OSM map by adding Gaussian noise with the corresponding standard deviation. 
This synthetic dataset allows us to systematically assess our method’s performance across different levels of positional uncertainty.

\begin{table}[tb]
    \centering
    \begin{sc}
    \caption{The statistics of real-world map-to-map matching datasets.}
    \begin{tabular}{lcccc}
    \toprule
         & \ \#nodes & \#edges & area \\
    \midrule
    Boston  & 2251 & 2438 & 2.2km $\times$ 1.8km   \\
    Ichikawa & 2496 & 2825 & 2km $\times$ 2km \\
    Shanghai & 161 & 178 & 0.6km $\times$ 0.5km  \\
    Boston-L & 10516 & 11821 & 5.2km $\times$ 2.7km \\
    Bremen & 119445 & 126999 & 23.67km $\times$ 32.96km\\
    \bottomrule
    \end{tabular}
    \label{tab:datasets}
    \end{sc}
\end{table}

\paragraph{Baseline Methods.} As the first method for map-to-map matching, we compare our approach against several relevant baseline methods. Since map data consists primarily of coordinate points, which are conceptually similar to point cloud data, we first compare our method to point cloud matching algorithms, including ICP~\cite{besl1992method} and SD~\cite{cuturi2013sinkhorn}. Additionally, we convert one of the maps into trajectory data by performing both Depth-First Search (DFS) traversal and Random Walk (RW) to simulate the movement of a vehicle along the roads.
Then these sampled segments are fed into a trajectory-map matching method HMM~\cite{newson2009hidden} for map-to-map matching. Furthermore, graph matching typically requires node-level supervision for training. To simulate this, we extract another group of data in OSM and introduce low, medium, and high noise perturbations to generate matching pairs, thus creating the necessary training labels for graph matching. We select the Neural Graph Matching (NGM)~\cite{wang2021neural} method as the representative graph matching approach. 

\paragraph{Evaluation.} For the map-to-map matching problem, we evaluate the performance based on the node matching results, where the correctness of the node correspondences directly influences the accuracy of road correspondences. Specifically, for each matched node, we identify the corresponding road in the other map and check whether the road matches. The Accuracy is calculated as the proportion of correctly matched roads. There are a few special cases in our evaluation to be treated differently. At intersections, a node may belong to multiple roads. In such cases, we consider the match correct if at least one of the corresponding road pairs is correct. Besides, if certain roads from the source map do not have corresponding roads in the target map, the nodes on these roads may be matched incorrectly, but this is not penalized as mismatch in the evaluation. This approach ensures that the evaluation is both realistic and fair, accounting for the complexity of real-world road networks and the inherent ambiguities in map data.


\subsection{Implementation Details}
\paragraph{Hyperparameters.} 
In our method, we use two neural networks: the first $\mathbf{\Psi}_{\theta}$ is an 3-layer Graph Convolutional Network (GCN)~\cite{kipf2016semi}, which utilizes 32 hidden dimensions. The second $\mathbf{\psi}_{\theta}$ is a two layer MLP with 32 hidden dimensions. For the Sinkhorn layer, we set the number of iterations to 20 to allow for sufficient optimization. The model is trained using the Adam ~\cite{kingma2014adam} optimizer with a learning rate of 0.001 for 200 epoches. These hyperparameter settings were chosen to balance model performance and computational efficiency, allowing for effective training and convergence across the evaluated tasks. 

\paragraph{Model Configuration} The experiments are conducted using an AMD EPYC 7542 CPU and a single NVIDIA 3090 GPU. For HMM method, the DFS order simulates the movement of a vehicle along the roads, converting the map data into a trajectory, which is used as a comparison against trajectory matching methods. Ramdom Walk starts from a random initial node, with the vehicle randomly selecting one of its neighboring nodes at each step. The process continues until all the neighbors of a node have been visited by the trajectory. Once this condition is met, the walk selects a new starting point from the unvisited nodes randomly and repeats the process. This continues until all nodes in the graph have been visited and converted into the trajectory. 


For the training of NGM, we select a region in Shanghai from the OSM data, covering an area of 68 km $\times$ 66 km. The map is segmented into 60 blocks for generating the training set. Each block is treated as an independent sample, and further divided into a $3 \times 3$ grid of overlapping tiles following the approach in Section~\ref{sec: discussion}. For each tile, we perform node shuffling to introduce structural variation and prevent overfitting to specific spatial layouts. This ensures that the model learns robust representations of road networks rather than memorizing fixed spatial patterns. Following node shuffling, we purturb them with low, medium, and high noise as what we do in our synthetic datasets. The ground truth matching pairs are generated according to the shuffle mapping, providing a supervised training signal for the model. For the GNN backbone, the same 3-layer GCN architecture used in our method is adopted. we train the NGM model for 200 epochs with a learning rate of 0.001 and evaluate it on three real-world datasets. Additionally, for both training and testing stages, we employ the same pseudo coordinates construction, Sinkhorn normalization, and Hungarian matching strategy ensuring a fair and consistent comparison between NGM and our proposed method.


The other baseline methods were adapted from their official source code repositories to ensure consistency with their original implementations.

\subsection{Results on Real World Datasets}
Experimental results on real-world datasets are summarized in Table~\ref{tab: acc real}. 
Our method consistently outperforms all baseline algorithms on the three datasets, achieving significant improvements in matching accuracy. This highlights the effectiveness of our approach in addressing the challenges of map-to-map matching. Meanwhile, we record the running time in Table~\ref{tab:runtime}, which shows that our method demonstrates competitive runtime performance, with execution times comparable to the fastest baseline methods.  The combination of high accuracy and low runtime makes our method suitable for real-world, large-scale applications.

\begin{table}[tbp]
    \centering
    \begin{sc}
    \caption{Results of matching Accuracy $(\%)$ $\uparrow$ on real-world datasets.}
    \label{tab: acc real}
    \begin{tabular}{lcccc}
    \toprule
    Dataset & Boston & Ichikawa & Shanghai & Bremen \\ \midrule
    ICP & 94.69 & 82.13 & 84.47 & 67.37\\
    SD & 94.94 & 81.46 & 79.50 & 67.12\\
    HMM + DFS & 81.16 & 78.76 & 61.49 & 27.32\\  
    HMM + RW & 37.41 & 60.92 & 48.45 & 23.15\\
    NGM & 85.96 & 68.83 & 80.12 & 44.85\\
    $\mathrm{UM}\textsuperscript{3}$ & \textbf{97.38} & \textbf{87.60} & \textbf{91.82} & \textbf{87.52}\\
    \bottomrule
    \end{tabular}
    \end{sc}
\end{table}



Point cloud matching methods perform better than other counterparts due to their emphasis on geometric spatial relationships. However, they overlook crucial graph structures like road connectivity. Our method outperforms existing approaches by integrating both geometric and structural information. While supervised graph matching techniques like NGM require extensive training and struggle with regional variations, our unsupervised approach offers superior adaptability without requiring labeled data.

\begin{figure}[tbp]\small
    \centering
    \resizebox{0.5\textwidth}{!}
    {
    \begin{tabular}{cc}
    \rotatebox{90}{\qquad \qquad \qquad Boston} & \includegraphics[width=1\linewidth]{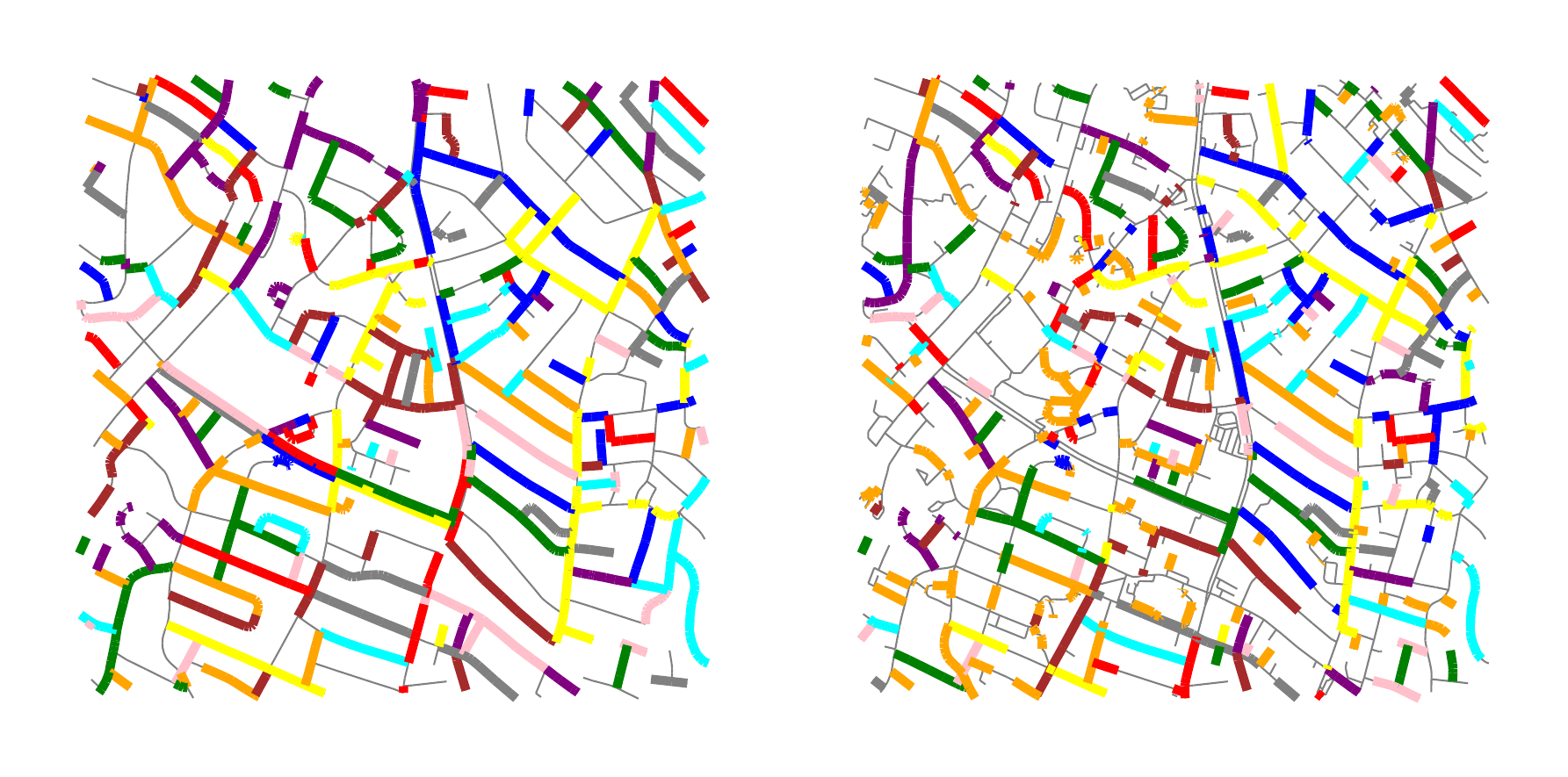} \\
    \rotatebox{90}{\qquad \qquad \qquad Ichikawa} & \includegraphics[width=1\linewidth]{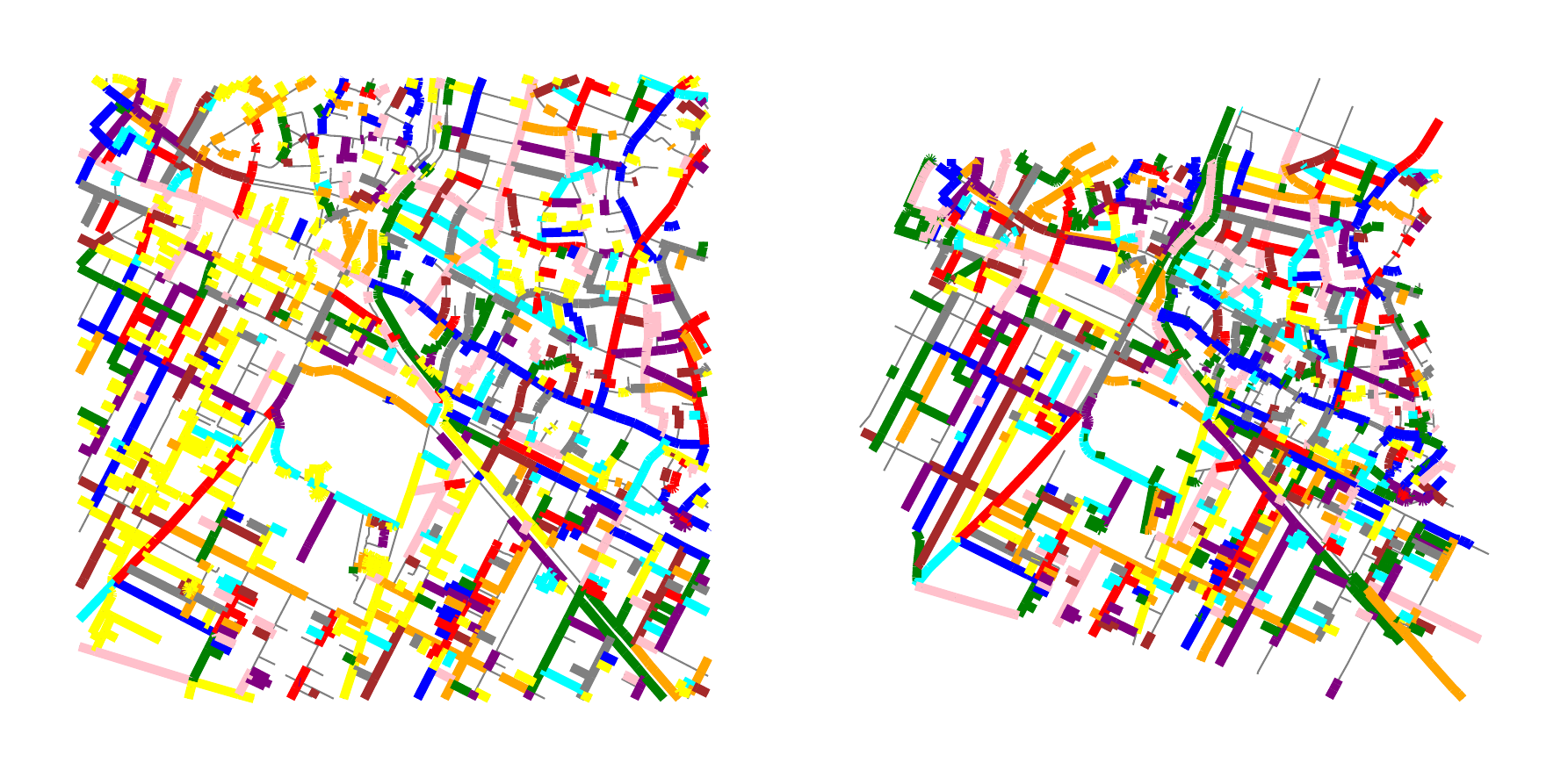}\\
    \rotatebox{90}{\qquad \qquad \qquad Shanghai} & \includegraphics[width=1\linewidth]{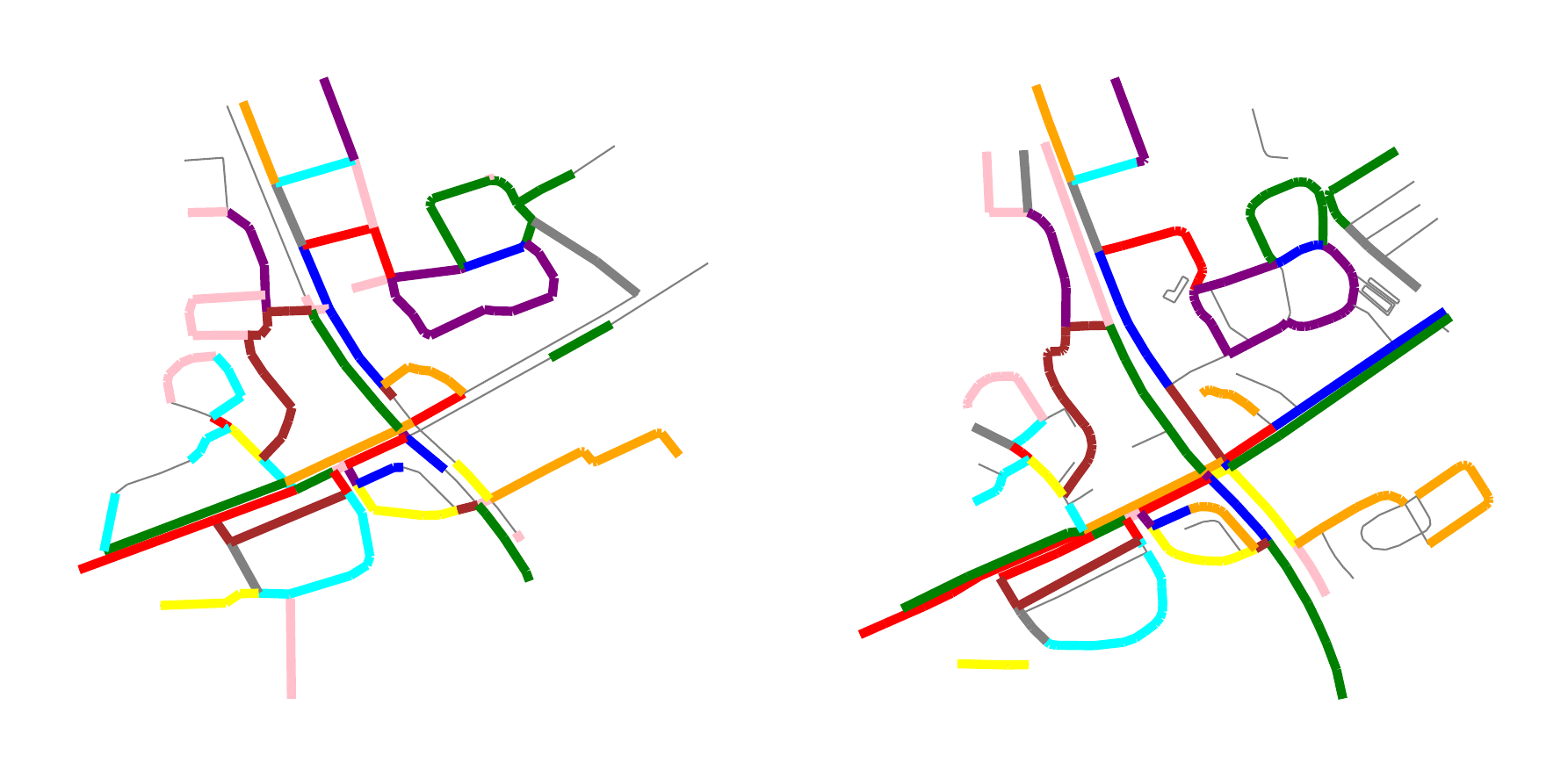}
    \end{tabular}
    }
    \caption{Visualization results of our method on real world datasets. Since the source and target maps are from the same region, the corresponding roads generally maintain similar relative positions across maps. The correctly matched roads are highlighted, while roads shown in thin black lines indicate unmatched or mismatched segments. Here, \textit{unmatched} refers to roads that have no corresponding counterparts in the other map (e.g., they do not exist in the other map).}
    \label{fig: visualization}
\end{figure}

The visualization results are presented in Figure~\ref{fig: visualization}. Since the map-to-map matching problem involves matching roads to other corresponding roads, we highlight the correctly matched roads. The roads that are either incorrectly matched or have no corresponding match in the ground truth are shown without highlighting. Each road is randomly assigned a color, and matching roads between the two maps are displayed using the same color to indicate their correspondence.

\begin{figure}[tbp]
    \centering
    \includegraphics[width=1.0\linewidth]{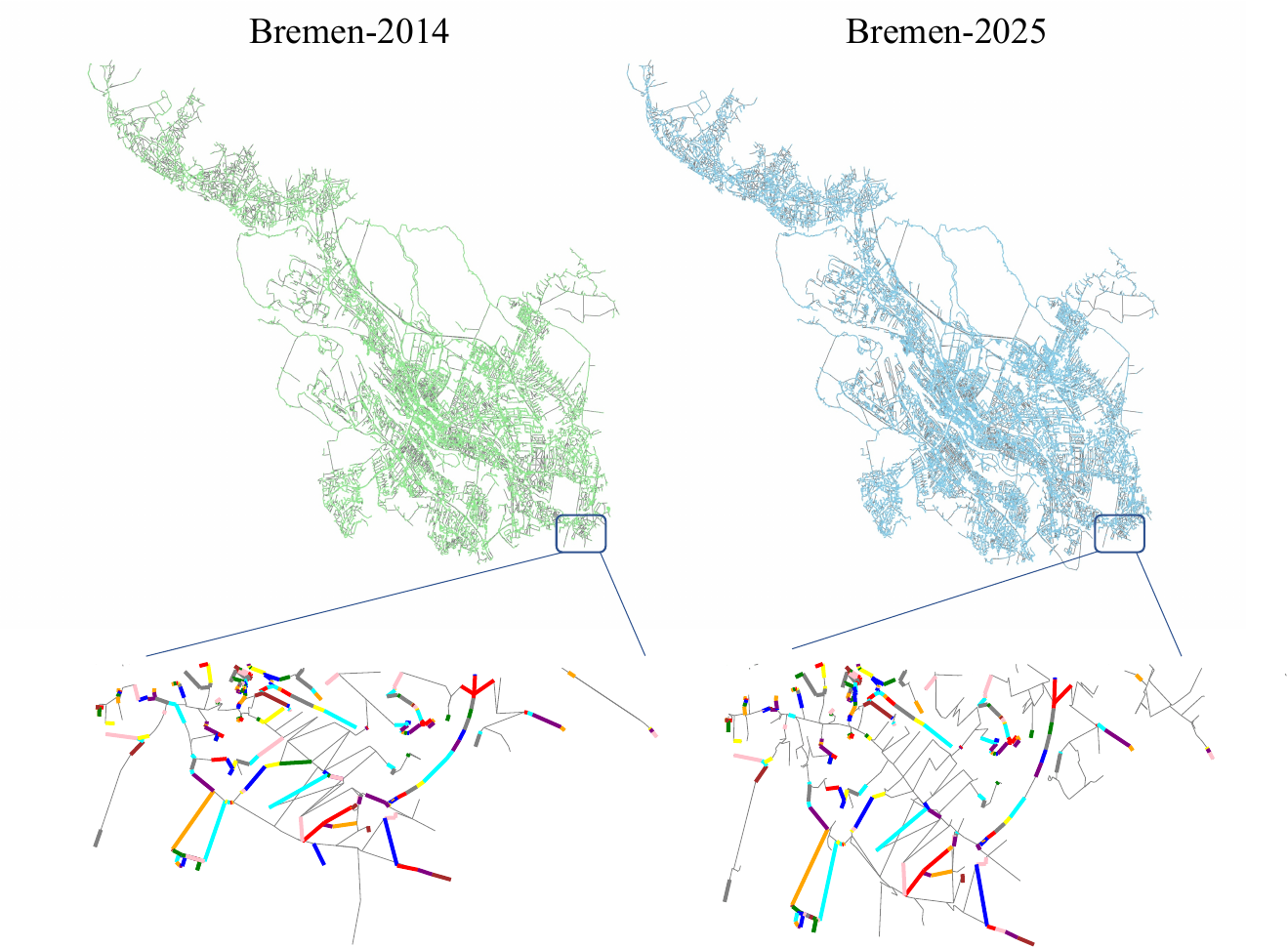}
    \caption{Node matching result of our method on Bremen dataset. The source and target maps were collected in 2014 and 2025, respectively. A zoomed-in view of a portion of the matching results is provided as an illustrative example.}
    \label{fig: bremen match}
\end{figure}

\noindent\textbf{On Large-scale Maps.} The Bremen dataset is considerably larger than the other datasets, which introduces more significant challenges for the map-to-map matching task. To this end, we adopt the tile-based strategy described in Section~\ref{sec: discussion} to fit the data on our device. Besides, although the source and target maps in this dataset come from the same location and the same source, significant differences have emerged over more than a decade of changes, making it still a challenging task to match the two maps. In particular, many roads present in the source maps no longer have corresponding counterparts in the target maps. These discrepancies are largely due to urban development, road reconstruction, and changes in mapping policies over time. From Table~\ref{tab: acc real}, we can observe that our method is still much more effective than the compared baselines. Due to the massive size of the maps, we visualize their overall structures and provide a zoomed-in view of a portion of the matching results as an illustrative example, as shown in Figure~\ref{fig: bremen match}.  The results indicate that our method remains effective in identifying regions with minimal changes and achieves high matching accuracy in these areas, largely due to the integration of both feature similarity and geometric similarity in our matching framework.

\begin{figure*}[tbp]
    \centering
    \includegraphics[width=.75\linewidth]{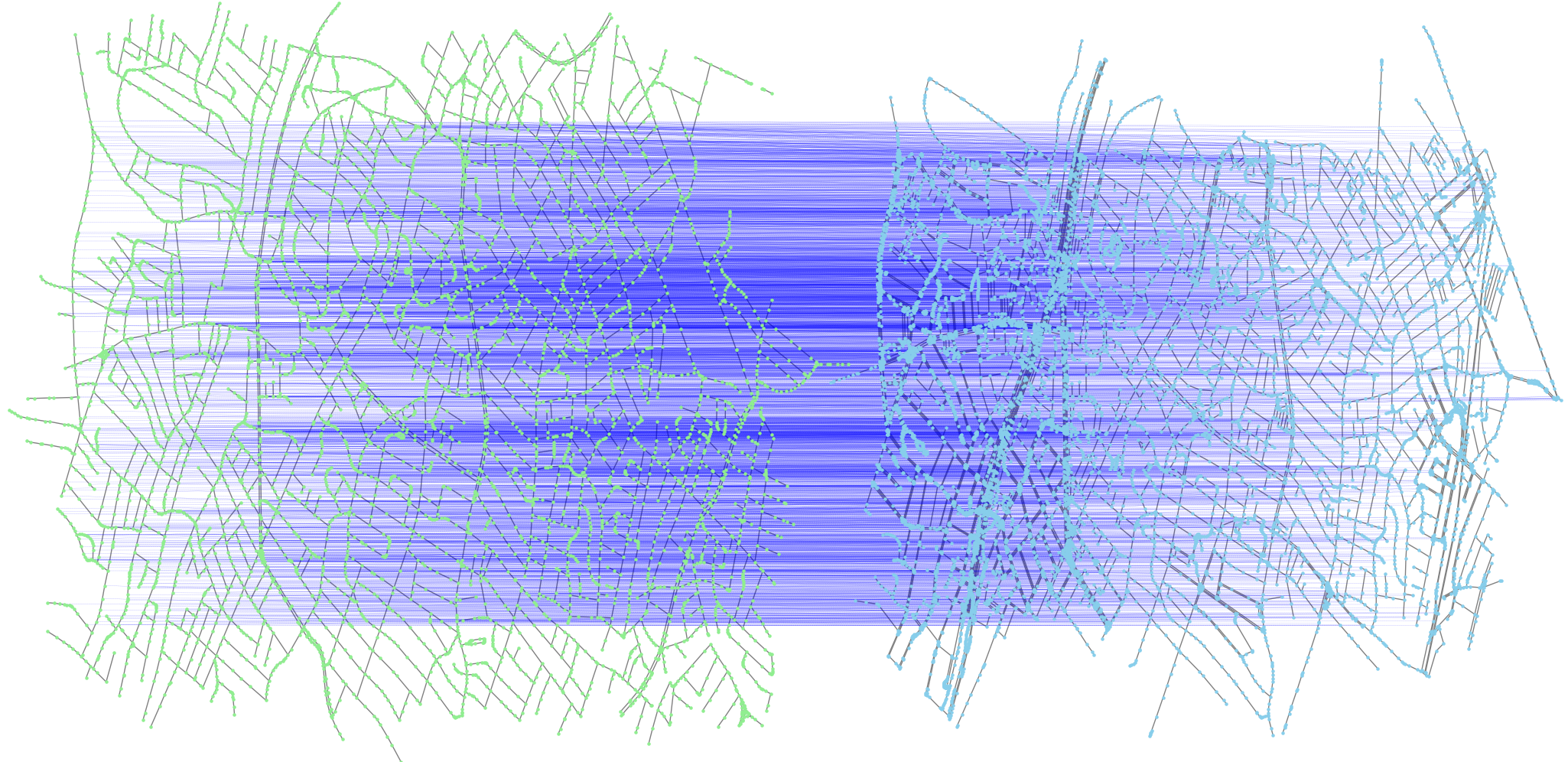}
    \caption{Node matching result of our method on Boston-L dataset. Each light blue line connects a pair of matched nodes from the two maps. Zoom in to better view.}
    \label{fig: boston match}
\end{figure*}

\noindent\textbf{On Maps with No Known Ground Truth.} We select a large map Boston-L and perform map matching against OSM data. Since the source and target maps come from different sources, it is not possible to directly obtain one-to-one correspondences as ground truth. Moreover, due to the large scale of the data, manual annotation is impractical, and thus we are unable to directly evaluate the performance of our method. Nevertheless, the quality of the map matching results can still be roughly assessed through visual inspection. As shown in Figure~\ref{fig: boston match}, the map matching results for the large-scale map remain highly accurate, further validating the scalability and reliability of our approach.


    

\begin{figure*}[tbp]
    \centering
    \subfigure[Varying $\alpha$]{
\includegraphics[width=0.3\textwidth]{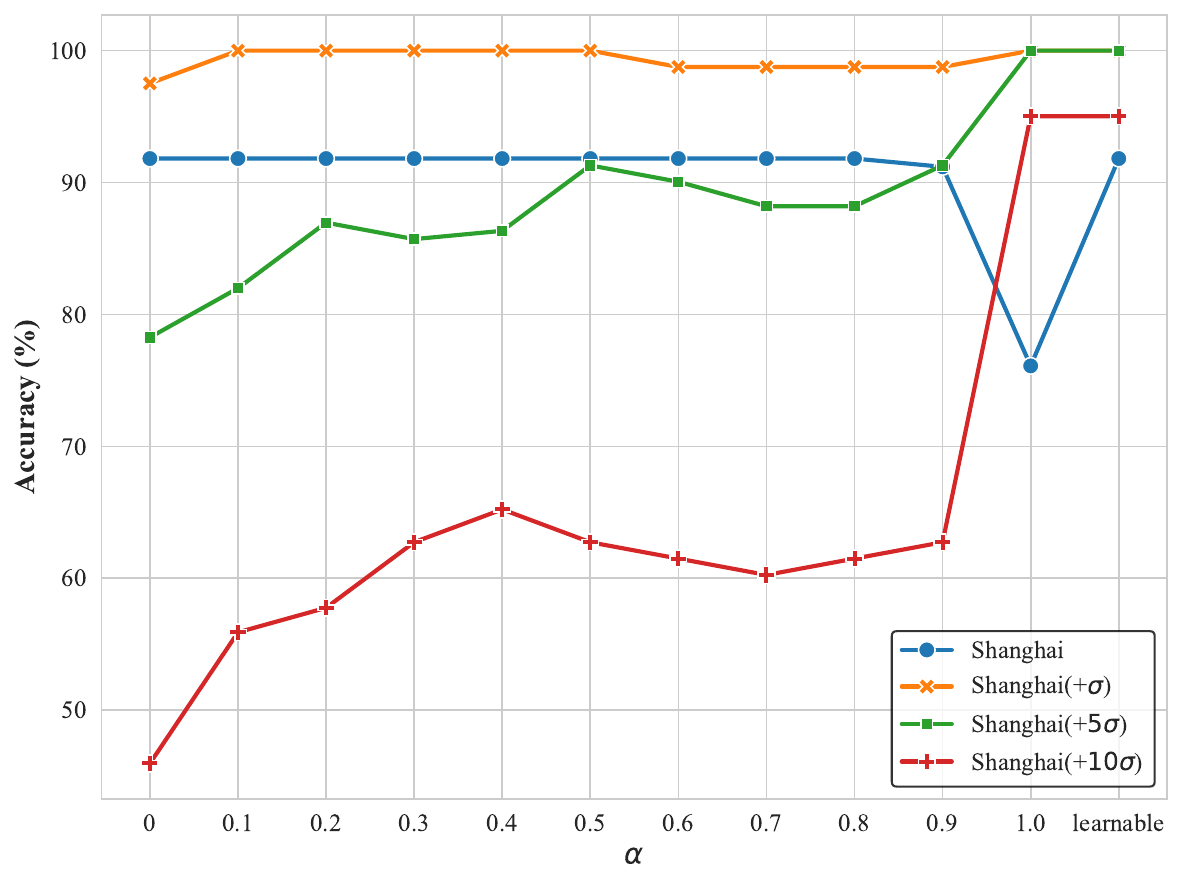}
         \label{fig:alpha}
    }
    \subfigure[Varying $\lambda$]{
 \includegraphics[width=0.3\textwidth]{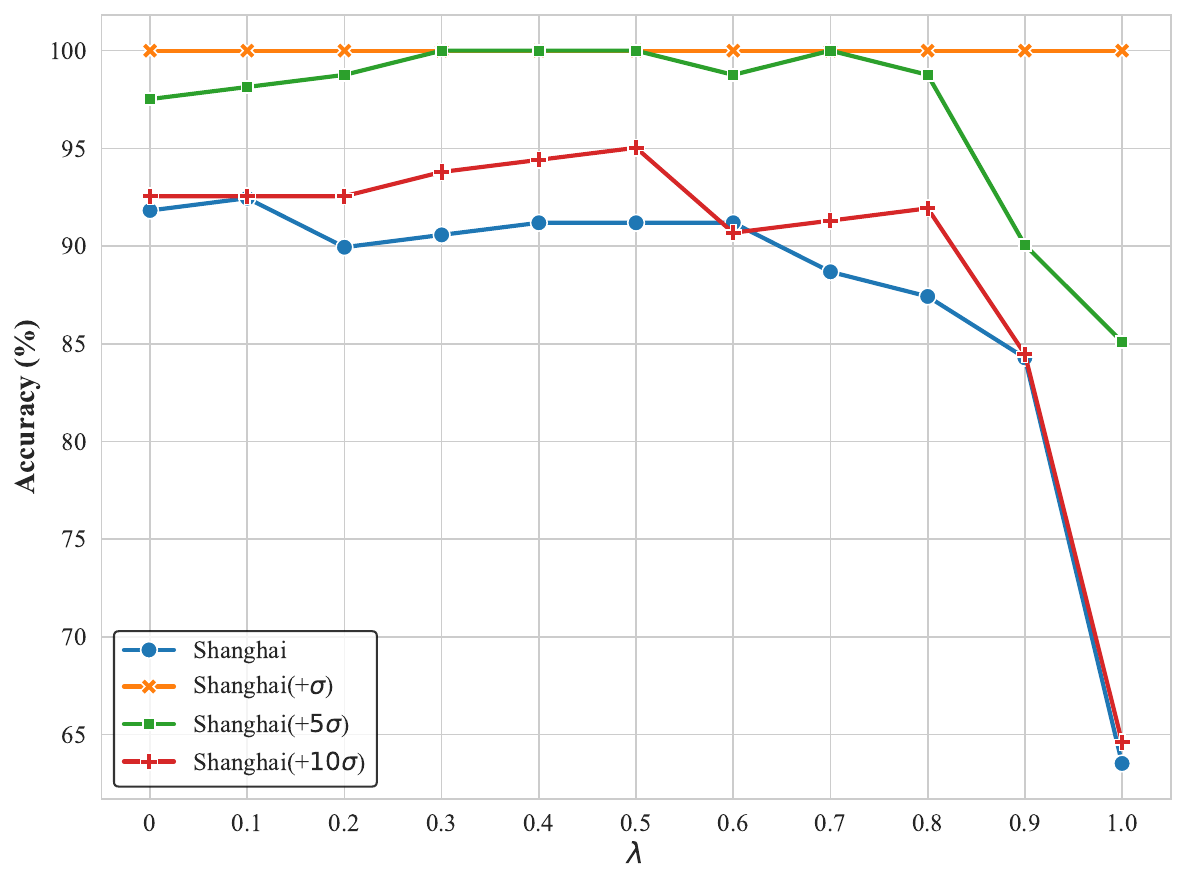}
        \label{fig:lambda}
    }
    \subfigure[Convergence analysis]{
 \includegraphics[width=0.3\textwidth]{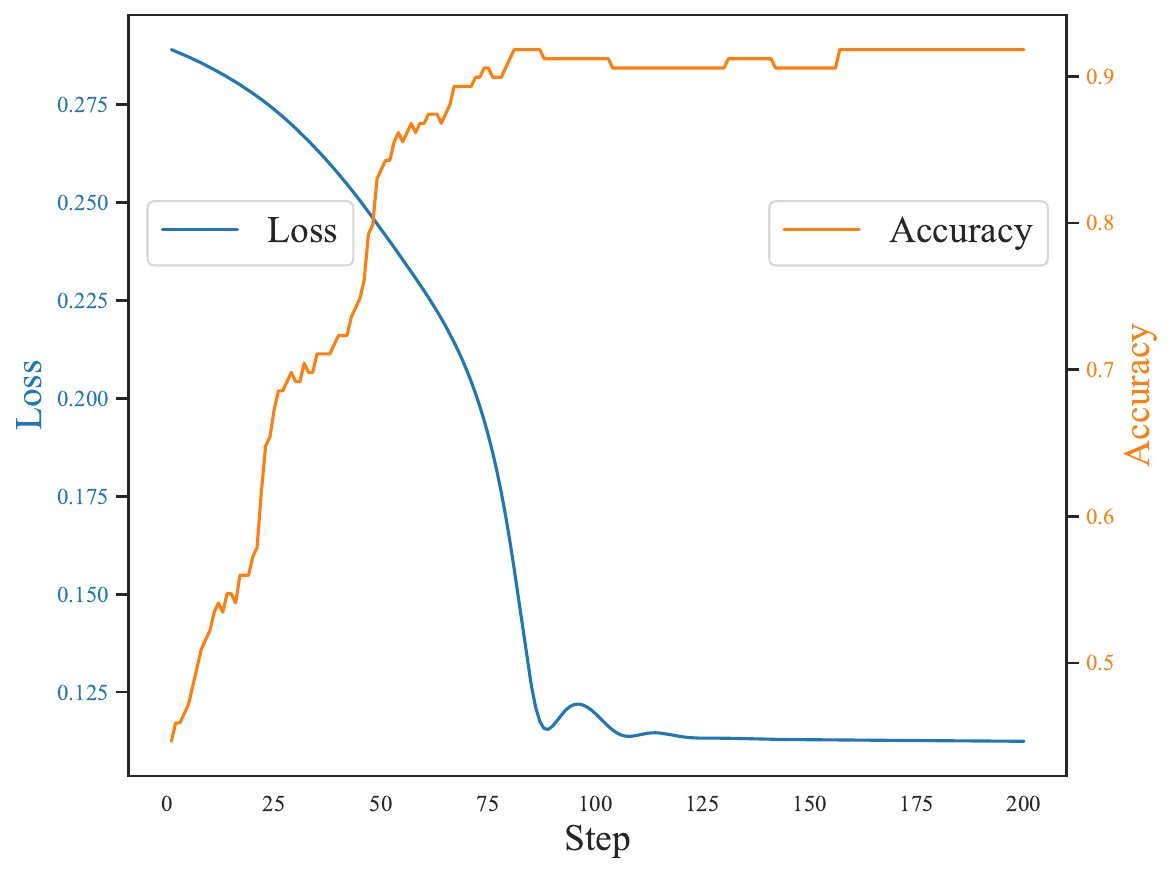}
        \label{fig:convergence}
    }
    \caption{Parameter analysis on the Shanghai dataset. (a) and (b) show the impact of varying parameters $\alpha$ and $\lambda$, respectively. The x-axis in both plots corresponds to different parameter values, and the y-axis denotes the matching accuracy. (c) illustrates the training dynamics, showing how the loss decreases and the accuracy improves over training steps. }
    \label{fig: ablation}
\end{figure*}




\subsection{Results on Synthetic Datasets}
To simulate real-world scenarios where map data can be highly complex, sourced from diverse origins, and often inaccurate, we evaluated our method and baselines on synthetic datasets with varying levels of noise. The results are presented in Table~\ref{tab: synthetic datasets}. Overall, our method demonstrates remarkable robustness to noise: it remains completely unaffected under low noise levels, nearly unaffected under medium noise, and only slightly impacted under high noise. This robustness highlights the practical applicability of our method in real-world scenarios, where data quality is often uncertain.

In contrast, the performance of other baseline methods degrades significantly even under low noise conditions. Under high noise levels, these methods almost entirely fail to provide reliable map-to-map matching results. Notably, the supervised learning method NGM, which was trained with extensive noisy data to enhance its robustness, performs more stably than other baselines. However, it still falls far short of the performance achieved by our method. These results underscore the superiority of our approach in handling noisy and imperfect data, further validating its potential for real-world applications.

\begin{table*}[tbp]
    \centering
    \begin{sc}
    \caption{Results of matching Accuracy $(\%)$ $\uparrow$ on synthetic datasets. We consider low, medium, and high noise scales to simulate real world noisy scenarios.}
    \label{tab: synthetic datasets}
    \begin{tabular}{lccc|ccc|ccc}
    \toprule
    \multirow{2}{*}{} & \multicolumn{3}{c|}{Low noise ($\sigma$)}   & \multicolumn{3}{c|}{Medium noise ($5\sigma$)} & \multicolumn{3}{c}{High noise ($10\sigma$)} \\ \cline{2-10}
                  & Boston          & Ichikawa          & Shanghai         & Boston         & Ichikawa        & Shanghai        & Boston          & Ichikawa          & Shanghai          \\
    \midrule
    ICP & 87.94 & 79.25 & 83.54 & 49.54 & 41.51 & 39.63 & 31.22 & 26.40 & 24.39\\
    SD & 87.88 & 79.12 & 76.52 & 50.83 & 41.88 & 39.02 & 32.31 & 25.86 & 20.73 \\
    HMM + DFS & 0.58 & 0.54 & 2.74 & 1.25 & 0.33 & 2.74 & 0.17 & 3.39 & 6.71\\
    HMM + RW & 1.12 & 0.54 & 2.13 & 1.12 & 1.84 & 3.66 & 1.08 & 0.38 & 3.66 \\
    NGM & 85.96 & 68.49 & 81.37 & 85.52 & 67.53 & 78.88 & 84.10 & 66.61 & 72.67\\
    $\mathrm{UM}\textsuperscript{3}$ & \textbf{100} & \textbf{100} & \textbf{100} & \textbf{99.82} & \textbf{99.76} & \textbf{100} & \textbf{95.82} & \textbf{97.64}& \textbf{95.03}\\
    \bottomrule
    \end{tabular}
    \end{sc}
\end{table*}

\subsection{Runtime Comparisons}
\label{appendix: runtime comparisons}
The runtime comparisons are shown in Table~\ref{tab:runtime}. In addition to achieving the best matching accuracy, our method demonstrates competitive runtime, with execution times comparable to the fastest baseline methods. This efficiency is achieved through our unsupervised learning framework, which iteratively refines the matching results without requiring expensive training or extensive computational resources. The combination of high accuracy and low runtime makes our method suitable for real-world, large-scale applications.

\begin{table}[tbp]
    \centering
    \begin{sc}
    \caption{Runtime comparisons (in seconds) on real world map-to-map matching datasets.}
    \label{tab:runtime}
    \begin{tabular}{lcccc}
    \toprule
    Dataset & Boston & Ichikawa & Shanghai & Bremen\\ \midrule
    ICP & 15.03 & 15.82 & 6.03 & $7.26 \times 10^2$\\
    SD  & 5.37 & 5.49 & 0.67 & $1.31 \times 10^2$\\
    HMM + DFS & $1.77 \times 10^2$ & $1.34 \times 10^2$ & 12.23 & $3.89 \times 10^4$\\
    HMM + RW & $1.12 \times10^3 $ & $9.02 \times 10^2$ & 37.69 & $5.01 \times 10^4$\\  
    NGM & $2.03 \times 10^4$ & $2.03 \times 10^4$ & $2.03 \times 10^4$ & $2.33 \times 10^4$\\
    $\mathrm{UM}\textsuperscript{3}$ & 9.92 & 8.59 & 3.56 & $5.17 \times 10^2$\\
    \bottomrule
    \end{tabular}
    \end{sc}
\end{table}

\subsection{Analysis}
\label{sec: parameter analysis}
\paragraph{Parameter Analysis.} Our method involves two key hyperparameters $\alpha$ and $\lambda$. The parameter $\alpha$ controls the contribution of feature similarity and geometric similarity, while $\lambda$ balances the trade-off between $\mathcal{L}_{dis}$ and $\mathcal{L}_{struct}$. In this section, we analyze the impact of these parameters on our model using the Shanghai dataset as an example.

As shown in Figure~\ref{fig:alpha}, we evaluate the performance of our model with different values of $\alpha$. Notably, when $\alpha = 0$, the model relies entirely on geometric similarity; when $\alpha = 1$, the model relies entirely on feature similarity. On the selected datasets, where map coordinates are relatively accurate, geometric similarity can complement feature similarity by providing additional spatial information that feature-based methods might miss. As a result, our model achieves strong performance when $\alpha$ is small. 
On noisy datasets, however, 
As shown in the results, the learnable $\alpha$ adaptively learns an optimal balance between feature and geometric similarity, achieving the best performance. In practical applications, users can also manually adjust $\alpha$ based on their prior knowledge or specific requirements.

In Figure~\ref{fig:lambda}, we evaluate the performance of our model with different values of $\lambda$. The parameter $\lambda$ balances the contributions of two proposed loss terms. A well-chosen $\lambda$ ensures that the model effectively leverages both geometric and structural information. When coordinate data is accurate, the computed distances are reliable, and $\mathcal{L}_{dis}$ provides strong guidance for the model to learn better matching relationships. However, in the presence of significant noise, distance information becomes less accurate, and $\mathcal{L}_{struct}$ offers stable support by leveraging structural consistency. This demonstrates that an appropriate balance between these two losses is critical for achieving robust and accurate matching results.

\paragraph{Convergence Analysis.} To further analyze the convergence behavior of our method, we visualize the changes in both loss and accuracy throughout the training process, using the Shanghai dataset as a representative example. As shown in Figure~\ref{fig:convergence}, the loss decreases steadily while the accuracy consistently improves, indicating that the model is able to effectively learn meaningful correspondences over time. This trend validates the design of our proposed loss function, which serves as a reliable surrogate objective for the unsupervised map matching task. The close alignment between the loss reduction and accuracy improvement also demonstrates that the loss captures key aspects of matching quality. Moreover, the method exhibits rapid convergence within a relatively small number of training steps, underscoring both its optimization efficiency and practical effectiveness.

\subsection{Ablation Study}
\label{sec: ablation study}

To demonstrate the significance of the pseudo coordinates module, we conduct an ablation study by removing this component and directly using raw latitude and longitude as initial features. We name this variant as No Pseudo Coordinates (-NPC). The experimental results in Table~\ref{tab: ablation pseudo coordinates} show a significant drop in performance when pseudo coordinates are omitted. This is because raw latitude and longitude coordinates within a small region exhibit minimal variation, lacking the discriminative power needed for effective matching. Since both our loss function and distance matrix rely on coordinate information for computation, the absence of meaningful pseudo coordinates leads to large matching errors.

Another part of the ablation study focuses on removing the fusion of feature and geometric similarity by using only one of them to model the node correspondence. This is achieved by setting $\alpha = 0$ or 1. Similarly, the ablation of the structure-based loss is performed by setting $\lambda = 0$. These variants have already been analyzed in Section~\ref{sec: parameter analysis}, which demonstrates the effectiveness of our proposed modules in addressing the map-to-map matching problem.

\begin{table}[tbp]
    \centering
    \begin{sc}
    \caption{Results of Accuracy $(\%)$ $\uparrow$ about the ablation study of our proposed pseudo coordinates.}
    \label{tab: ablation pseudo coordinates}
    \begin{tabular}{lcc}
    \toprule
    & $\mathrm{UM}\textsuperscript{3}$-NPC & $\mathrm{UM}\textsuperscript{3}$ \\
    \midrule
    Shanghai & 31.45 & 91.82 \\
    Shanghai $+ \sigma$ & 98.76 & 100\\
    Shanghai $+5\sigma$ & 80.75 & 100\\
    Shanghai $+10\sigma$ & 49.69 & 95.03\\ 
    \bottomrule
    \end{tabular}
    \end{sc}
    \vspace{-2mm}
\end{table}

This experiment highlights the critical role of pseudo coordinates as a foundational component of our method. By generating meaningful and discriminative coordinate representations, the pseudo coordinates module effectively enhances the model's learning capability, ensuring robust and accurate map-to-map matching.







\section{Conclusion}
\label{sec: conclusion}

In this paper, we introduce the first unsupervised map-to-map matching method, addressing the challenge of aligning maps from different sources. Unlike traditional point cloud matching or trajectory-map matching approaches, our method explicitly models the structural relationships in maps while incorporating geometric similarity learning to improve robustness against noise. We evaluate our approach on several real-world and synthetic datasets from different regions, and compare it against competitive baselines. The results demonstrate that our method is both effective and efficient, especially in noisy scenarios. Our method also exhibits high scalability due to our designed tile-based multi-processing strategies.

Our work establishes a strong foundation for future research in map-to-map matching and opens up new possibilities for applications in map alignment, navigation, and geographic data integration. It demonstrates the feasibility of learning accurate correspondences without requiring manual annotations. In future work, we aim to explore more advanced similarity measures and investigate how our method can be extended to handle dynamic changes in road networks over time.


\bibliographystyle{ACM-Reference-Format}
\bibliography{sample-base.bib}





    

\end{document}